# Towards Retrieval Augmented Generation over Large Video Libraries

Yannis Tevissen
Moments Lab Research
Boulogne-Billancourt, France
yannis@momentslab.com

Khalil Guetari
Moments Lab Research
Boulogne-Billancourt, France
khalil@momentslab.com

Frédéric Petitpont
Moments Lab Research
Boulogne-Billancourt, France
fred@momentslab.com

**Abstract**— Video content creators need efficient tools to repurpose content, a task that often requires complex manual or automated searches. Crafting a new video from large video libraries remains a challenge. In this paper we introduce the task of Video Library Question Answering (VLQA) through an interoperable architecture that applies Retrieval Augmented Generation (RAG) to video libraries. We propose a system that uses large language models (LLMs) to generate search queries, retrieving relevant video moments indexed by speech and visual metadata. An answer generation module then integrates user queries with this metadata to produce responses with specific video timestamps. This approach shows promise in multimedia content retrieval, and AI-assisted video content creation.

## I. INTRODUCTION

Video content creators, journalists and media managers spend a significant amount of time searching for ways to repurpose their content to tell new stories. This work is usually done through manual search or automated search engines that are usable only by knowledgeable users. Finding the right content to illustrate a story requires search abilities but also some understanding of what would make a good story based on what users want to create. For archive content repurposing, it also requires a vast knowledge of the past.

AI-assisted content creation is becoming more and more frequent. When it comes to textual creations, LLMs can serve as an ideation assistant as well as a full writing aid following natural language instructions. For multimedia such experiments are often centered about fully synthetic contents such as images, music or videos. But for many content creation projects such as news journalism or documentaries, using synthetic content is simply impossible as the main goal is to illustrate and explain some real events. That is why building new stories from video moments stored inside large video libraries is key.

Recent advances in natural language processing have permitted the use of large language models for various tasks and especially as knowledgeable assistants when it comes to answering questions based on a large quantity of content [1]. These capabilities are well known under the name of retrieval augmented generation (RAG) [2] and are now used in a wide variety of domains [3] from legal question answering [4] to healthcare [5]. These methods essentially rely on RAG over textual database. It is particularly more complex to perform RAG over multimedia database and especially video libraries. Performing RAG over video contents requires several more core components specific to video data processing.

This work presents an architecture and some examples of retrieval augmented generation over large video libraries. This stands at the intersection of video moment retrieval and video question answering. It could be called video library question answering (VLQA).

The remainder of this paper is organized as follows: Section II reviews related work in video-text retrieval and RAG. Section III describes our proposed architecture in detail. Section IV presents our experimental setup and early results. Finally, section V discusses the advantages, limitations, and potential future directions.

## II. RELATED WORK

Video-text retrieval has been a major topic in computer vision for the past years [6] with the introduction of various techniques. The vast majority of them use image embeddings generated with models such as CLIP [7] coupled with temporal representations [8] to obtain vector representations of short videos. From uncertainty addition through learned parameters [9], to the use of diffusion models [10], recent video retrieval methods now achieve good performances on challenging benchmarks such as MSR-VTT [11].

Some works also focus on finding video moments [12], [13], [14] instead of full-length videos, setting a path towards video retrieval compatible with long-form videos. Some other methods also propose to use these local information to improve retrieval thanks to embeddings created at different granularity levels [15].

If applying RAG to videos is not an easy task, iRag [16] still proposes a method for understanding long form videos with RAG applied over extracted metadata.

Finally, LLMs have also been used as agents [17] capable of using specific tools to perform a set of predefined tasks. This has been used for long form video understanding to achieve competitive results on video question answering benchmarks [18] and also for video editing [19].

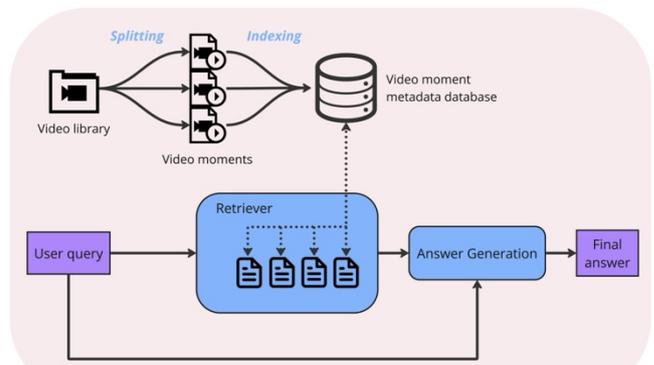

Fig. 1. Overview of the proposed architecture for retrieval augmented generation over large video libraries.

## III. ARCHITECTURE

We introduce a simple architecture for video moment retrieval within large video libraries. This architecture illustrated in Fig. 1 relies on two main components: a retriever module and a conversational module to generate the final answer.

### A. Retriever module

The role of the retriever module is to generate queries that can be sent to a text-based search engine such as OpenSearch [20]. In this case, a LLM is prompted to generate at least 5 search queries composed of a few keywords. Each search query outputs some video moments that are represented as text documents with the following format. Up to 50 documents are kept and transmitted to the answer generation module. This number can vary to ensure the desired equilibrium between precision and rapidity of answer. The more moments the system try to recover in this phase, the longer will be the final LLM call (cf. III. B.). Finally, depending on the quality of the moment database and the search engine, there is a high risk of including irrelevant moments.

*1) Large language model*

Every large language model finetuned for instruction answering can be used in this architecture as the model is not being asked to generate any specific pattern or use any tools. The "smarter" the model, the better the analysis of the retrieved moments will be. When dealing with archive videos, carrying a good knowledge of human history can also be a great advantage especially to maximize the counterfactual robustness of the model [21]. Our early results showed particularly good performances with Mistral Large, Mixtral 8x7B [22] and Command-R [23]. The first one was used in Fig 2. examples.

*2) Video database and indexing strategy*

For this architecture to work, the video moments should be stored within a database that allows for an efficient video moment retrieval. It requires a splitting and indexing strategy to create small-enough video moments.

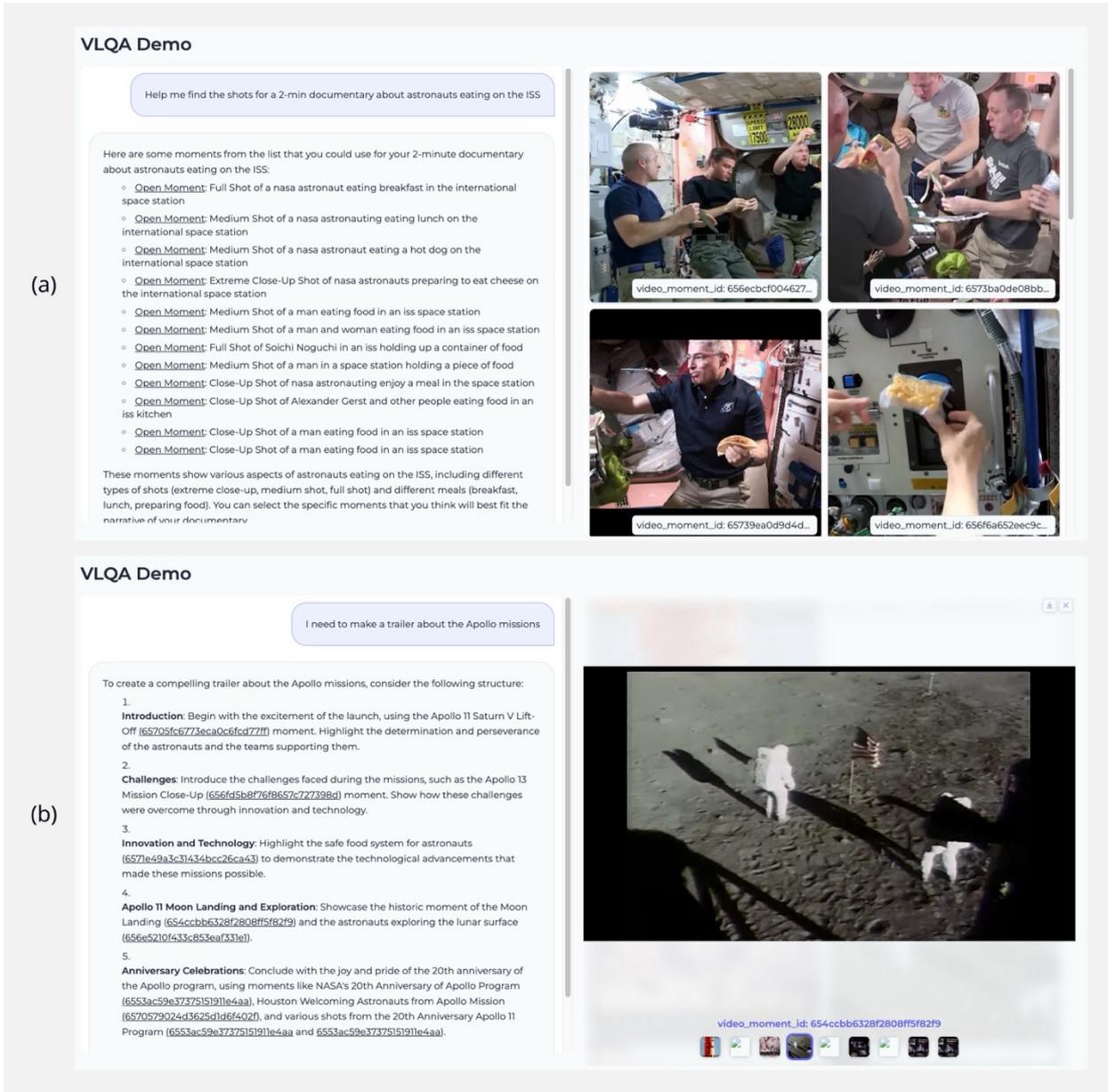

Fig. 2. Two conversationnal results and moments overviews obtained with our architecture. (a) illustrates the answer to the user query "Help me find the shots for a 2-min documentary about astronauts eating on the ISS" and (b) the answer to "I need to make a trailer about the Apollo missions". In both cases the answer generated include references to relevant video moments from a very large moments database.

Prior to metadata extraction, all videos were split into video moments. We used a content-aware video splitting algorithm [24] that splits the video based on HSV pixel color variations.

For indexing, several options coexist to create video embeddings [25], [26], [27] but for this study we preferred a metadata-based approach that combines several expert systems into a large document database. This approach has already proven to be particularly competitive with state-of-the-art VideoLLMs [16], [28], [29], [30], [31] especially when the textual metadata extracted are then summarized by a LLM [32]. For this work the moment indexing was done using pyannote speaker diarization [33] and whisper [34] to transcribe what was said. We also extracted for each video clip 3 uniformly distributed frames and captioned them with BLIP2 [35]. Together, these are the metadata that populated the video moment metadata database.

*B. Answer generation*

Finally, an answer generation module gathers the initial user query and the relevant video moments metadata found by the retriever into a larger prompt that is given to a LLM to generate the final answer. In our case the LLM is prompted to include references to the video moments in a specific format: *[video_id](timestamp_in;timestamp_out)*. This enables the system to further replace these references with hyperlinks to the video moments hosted online.

## IV. EARLY RESULTS

Experiments were run on a very large video library extracted from NASA assets freely available for educational purposes. The full library used contains 9388 video files with very varying durations from 10 seconds to more than 2 hours. The videos have been split into 372181 video moments that were then indexed to constitute our final video moment metadata database.

Some results are presented in Fig. 2 highlighting two functional use cases of this architecture. First the proposed architecture can answer conversational text queries in the wild while making references to relevant video moments. Secondly, it allows for complex searches of footage to illustrate a concept such as "the Apollo missions". Such search queries were previously difficult to get right but this method can efficiently find both relevant moments and propose a trailer script accordingly to the moments found.

In the examples given, both the responses took about 15 seconds to be generated.

## V. DISCUSSION

*A. Advantages of the method*

One of the key advantages of the method is its ability to retrieve specific moments, even when they contain no speech. This is permitted using the image captioning model that enrich the moment metadata with visual descriptions.

This method is also highly interoperable as it is agnostic of all the elements used. It can work with a wide variety of LLMs and can be adapted with all kinds of text-based search engines. It is also fast as the search queries can be run in a few milliseconds and the LLMs are generating short answers from relatively short prompts.

Even with recent advances towards vLLMs capable of handling millions of token [36], handling several hour-long videos remains a challenge. Performing RAG over large video moments databases does not rely on particularly long context window handling of the LLM and therefore enables question answering over thousands of hours of videos.

*B. Limitations*

This method cannot be used to generate detailed answers about a video library other than what was included in the generated metadata. It relies on carefully chosen metadata indexing. Also, it cannot perform certain actions such as counting event or object occurrences in the full video library, neither can it provide precise insights on the duration of appearance of things.

Hallucinations are also an issue. Apart from standard hallucinations that occur for every LLM, it becomes especially problematic with LLMs trained on large unfiltered web databases such as FineWeb [37] because they contain a lot of hyperlinks to existing videos on the web. Therefore, it becomes quite frequent that the LLM hallucinates a link to a YouTube video instead of using the video moments provided by the retriever module.

*C. Need for benchmarks*

Although this paper showed some promising results of retrieval augmented generation over large video libraries, evaluating the performances of such systems remains a challenge. Several metrics exist to evaluate RAG methods over large text libraries. Most of them are summarized on MTEB leaderboard [38] but none include an element that could provide a clue on the ability of LLM to choose a good video segment based on a retrieved description.

Similar to Video Question Answering (VQA) benchmarks [11], [39], and VideoLLMs evaluation suites [40], [41], [42], it is necessary to build a standardized benchmark for questions over large video libraries.

*D. Future works*

Evaluating Video Library Question Answering (VLQA) also requires a proper dataset made of several video libraries, questions and their expected answer built with video moments from the library.

Apart from creating a dataset and benchmark to evaluate such methods, future works to improve this architecture could include the addition of a reranker [43] module trained specifically to choose the correct video moments to answer the user query.

## VI. CONCLUSION

In this paper we demonstrated the possibility to apply retrieval augmented generation to large video libraries. We also introduced a new task called Video Library Question Answering (VLQA) that requires both language and video understanding capabilities. Future works will focus on creating benchmarks for this task and incorporating a multimodal reranker module inside our architecture. Overall, this study highlights the potential of RAG for enhancing multimedia content utilization with new creative workflows and AI-assisted creation based on real footages.